\documentclass[10pt,twocolumn,letterpaper]{article}

\usepackage{iccv}
\usepackage{times}
\usepackage{epsfig}
\usepackage{graphicx}
\usepackage{amsmath}
\usepackage{amssymb}
\usepackage{color}
\usepackage{soul}

\usepackage{booktabs}       %
\usepackage[accsupp]{axessibility}  

\usepackage[pagebackref=true,breaklinks=true,letterpaper=true,colorlinks,bookmarks=false]{hyperref}

\iccvfinalcopy 

\def\iccvPaperID{****} 

\def\R{{\mathbb R}}
\def\ie{\emph{i.e}\onedot} 
\def\ece{ECE}

\def\enet{ECENet}

\ificcvfinal\fi

\begin{document}

\title{Boosting Semantic Segmentation from the Perspective of\\Explicit Class Embeddings}

\author{
    Yuhe Liu\textsuperscript{\rm 1,2}\thanks{Work done during an internship at Huawei Noah’s Ark Lab.}\quad\quad
	Chuanjian Liu\textsuperscript{\rm 2}\quad\quad
	Kai Han\textsuperscript{\rm 2\textdagger}\quad\quad
    Quan Tang\textsuperscript{\rm 3}\quad\quad
    Zengchang Qin\textsuperscript{\rm 1}\thanks{Corresponding author.}
	\and
    {\textsuperscript{\rm 1}Beihang University}\quad\quad
    {\textsuperscript{\rm 2}Huawei Noah’s Ark Lab}\quad\quad
	{\textsuperscript{\rm 3}South China University of Technology}
	\and
	{\tt\small \{liuyuhe, zcqin\}@buaa.edu.cn \quad\quad \{liuchuanjian, kai.han\}@huawei.com}
	\and
	{\tt\small csquantang@mail.scut.edu.cn}
}

\maketitle
\ificcvfinal\thispagestyle{empty}\fi

\begin{abstract}
	Semantic segmentation is a computer vision task that associates a label with each pixel in an image. Modern approaches tend to introduce class embeddings into semantic segmentation for deeply utilizing category semantics, and regard supervised class masks as final predictions. In this paper, we explore the mechanism of class embeddings and have an insight that more explicit and meaningful class embeddings can be generated based on class masks purposely. Following this observation, we propose ECENet, a new segmentation paradigm, in which class embeddings are obtained and enhanced explicitly during interacting with multi-stage image features. Based on this, we revisit the traditional decoding process and explore inverted information flow between segmentation masks and class embeddings. Furthermore, to ensure the discriminability and informativity of features from backbone, we propose a Feature Reconstruction module, which combines intrinsic and diverse branches together to ensure the concurrence of diversity and redundancy in features. Experiments show that our ECENet outperforms its counterparts on the ADE20K dataset with much less computational cost and achieves new state-of-the-art results on PASCAL-Context dataset. The code will be released at \url{https://gitee.com/mindspore/models} and \url{https://github.com/Carol-lyh/ECENet}.

\end{abstract}

\section{Introduction}

Semantic segmentation is a fundamental task in computer vision, which aims to 
predict the corresponding classes for each pixel of the input 
image. Typically, pixels that share common semantic categories are aggregated 
together to form regions on each slice of predicted masks, which 
naturally presents the description of each category the model has learned.

Traditional semantic segmentation methods are dominated by 
Fully Convolutional Networks (FCN)~\cite{long2015fully} based models. With 
stacked convolutional layers, the semantics in input images are gradually 
extracted. The $1\times1$ convolutional layer, which serves as 
semantic kernels, is
usually applied to the representative feature maps in the end. Previous 
works~\cite{chen2017deeplab, deeplabv317, fu2018dual, lin2017feature} focus on 
enlarging receptive filed~\cite{deeplab15, chen2017deeplab, deeplabv317}, integrating 
attention modules~\cite{fu2018dual, yuan2018ocnet, huang2019ccnet} or fusing 
multi-stage features~\cite{pinheiro2016learning, lin2017feature, 
kirillov2019panoptic}. However, the CNN architectures are lack 
of long-range dependencies, which hinders the performances of FCNs.

Recently, transformer~\cite{vaswani2017attention} using 
self-attention mechanism is introduced into the field of computer vision. 
Inspired by Vision Transformer (ViT)~\cite{dosovitskiy2020vit} and 
Segmenter~\cite{strudel2021segmenter}, class tokens/embeddings have 
aroused the interest of many
researchers~\cite{lin2022structtoken, cheng2021per, cheng2022masked, 
zhang2021k}. Generally, class embeddings are \textbf{randomly initialized} and passed into the decoder to interact with feature maps, then it would 
be used to get final segmentation masks~\cite{cheng2021per, cheng2022masked}. 
However, the class embeddings here 
are
defined implicit and meaningless initially, which means that much spatial \textbf{prior knowledge} is ignored and lost.

\begin{figure}[t]
	\centering
	\includegraphics[width=0.97\linewidth]{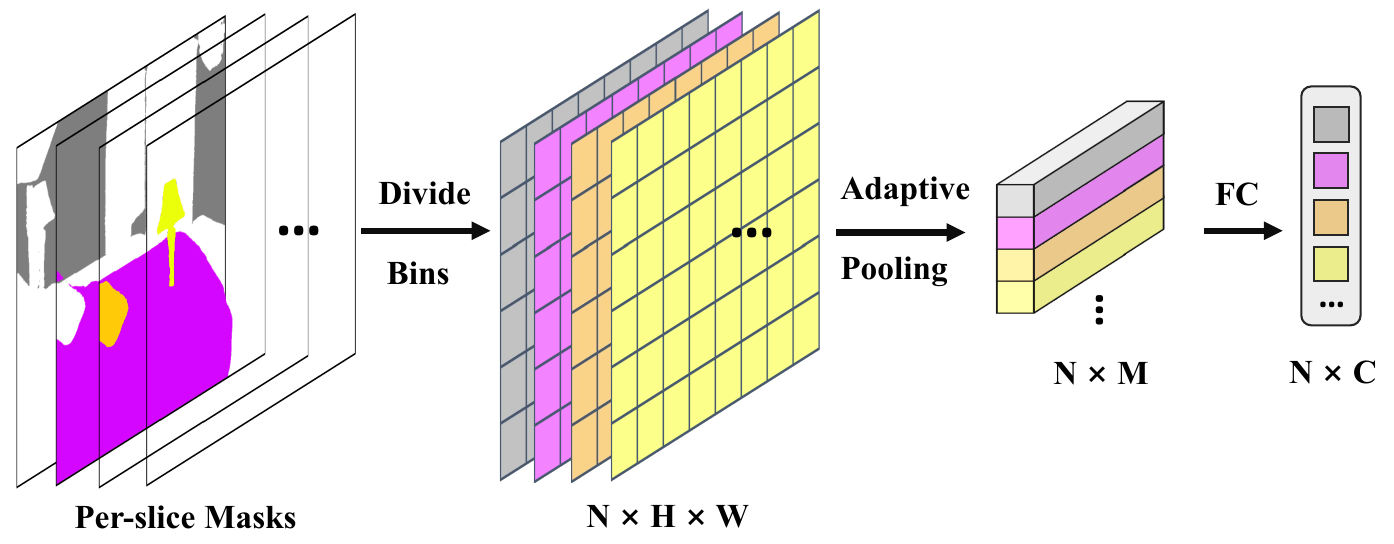}
	\caption{Explicit Class Extraction module. Each slice of predicted masks naturally presents what the model has learned on this category. By extracting class information into embeddings, we allow the information flow reversely and obtain the explicit class embeddings upon spatial prior knowledge on each region.}
	\label{fig1}
\end{figure}

Our key insight: class embeddings can be made \textbf{explicit} and \textbf{meaningful} by taking use of predicted masks. Intuitively, accurate regions on each slice of masks 
which the model has learned become the most natural description of each category.
In fact, MaskFormer~\cite{cheng2021per} first realizes the importance of per mask and replaces this per-pixel classification task with a set of binary masks prediction, each associated with a single category. Class embeddings which serve as object queries are passed into transformer decoder and then used to get the predicts. Given this promising attempt, a natural question emerges: \emph{can we reverse this process? Or can predicted masks assist the generation of meaningful class embeddings conversely?}

To address this question, we consider a simple approach to utilize the predicted masks in turn and propose our ECENet, which is composed of Feature Reconstruction (FR), Explicit Class Extraction (\ece), Semantics Attention \& Updater (SAU). More specifically, we design a FR module which is used to ensure the discriminative and informative capability of backbone features. Then Explicit Class Extraction (ECE) module consisting of spatial pooling and a single linear projection is used to extract class embeddings from masks. The class embeddings extracted are then passed into a transformer block that takes the early features as queries and class embeddings as keys and values. Through this, we encourage the interaction between early stage features and meaningful class semantics. Inspired by SegViT~\cite{zhang2022segvit}, we utilize the similarity maps as an extra masks. Then we employ gated mechanism~\cite{chung2014empirical} to further enhance previous class embeddings with the newly obtained semantics from the masks, dubbed Semantics Attention \& Updater (SAU). Since the masks are a byproduct of the regular attention calculations, negligible computation is involved. We get the final segmentation masks by aggregating the enhanced outputs from multi-stage features and employ a convolutional layer onto it, assisted by class embeddings and intermediate masks.

Building upon this effective paradigm, the regions belonging to the same category tend to group together with the assistance of highly consistent semantics. Furthermore, the semantic gap between different layers is bridged. Experiments show that our ECENet achieves promising performance on common segmentation datasets and outperforms its counterparts with less computational cost.

We summarize our main contributions as follows:
\begin{itemize}
	\item We reverse the general decoding process which departs from randomly initialized embeddings. Our class embeddings are explicit and consistently meaningful. It is the first attempt to uncover the correlations between segmentation masks and class embeddings and explore possible inverted information flow between both. 
	
	\item We propose a new network, ECENet based on this insight, which is composed of Feature Reconstruction (FR), Explicit Class Extraction (\ece), Semantics Attention \& Updater (SAU), which is demonstrated to be effective and efficient.
	
	\item We conduct extensive experiments on challenging benchmarks. Results show that our ECENet achieves competitive mIoU $55.2\%$ on the ADE20K dataset with much less computational cost and parameters. We also demonstrate that our method yields \textbf{state-of-the-art} results on PASCAL-Context dataset ($65.9\%$ mIoU) and is compelling on Cityscapes dataset.
\end{itemize}

\section{Related work}

\subsection{Semantic segmentation}

Fully Convolutional Networks (FCN)~\cite{long2015fully} based models dominate the field of segmentation in early research. With stacked convolutional layers, the semantics in input images are gradually extracted. Meanwhile, the resolution of feature maps is reduced concerning computational cost and limited memory, which naturally forms hierarchical feature maps and broadens the receptive field. However, the inherent limited context information still hinders the performance of FCNs. To resolve this, many previous works focus on enlarging receptive filed~\cite{deeplab15, chen2017deeplab, deeplabv317} or integrating attention modules~\cite{fu2018dual, yuan2018ocnet, huang2019ccnet}.

DeepLab~\cite{deeplab15} and DeepLabV2~\cite{chen2017deeplab} expand the receptive field by using dilated convolution. Despite this, an alternative approach is to integrate attention modules~\cite{fu2018dual, yuan2018ocnet, huang2019ccnet}. SENet~\cite{hu2018squeeze} won the championship of the ImageNet 2017 image classification task by adding a channel attention mechanism to adjust the channel response adaptively. DANet~\cite{fu2018dual} proposes the double attention network to simultaneously capture the global dependence in both spatial and channel dimension. ~\cite{wang2021exploring} explores cross-image pixel contrast to focus on global context of the training data differently. Simultaneously, many approaches~\cite{huang2019ccnet, wang2020eca, fu2020scene, wang2020axial} have been proposed to reduce the computational cost while retaining global attention.

\subsection{Transformers for vision}

Recently, transformer~\cite{vaswani2017attention} architecture which can capture long-range dependencies has replaced CNNs as the new backbone to extract features. According to spatial size of the feature maps, transformer can be divided as plain~\cite{dosovitskiy2020vit} and hierarchical~\cite{liu2021swin, wang2021pyramid, xie2021segformer, ren2022shunted} architectures. While the former remains the same resolution for all layers and the latter generally employs patch-merge methods between stages to get hierarchical resolutions. 

Besides being used as a backbone, attention-based transformer structure is also designed as a decoder to extract high level semantic information. Segmenter~\cite{strudel2021segmenter}, K-Net~\cite{zhang2021k}, StructToken~\cite{lin2022structtoken} and OneFormer~\cite{jain2023oneformer} introduce learnable tokens or dynamic filters into Transformer decoder. Visual Parser~\cite{bai2021visual} emphasizes the part-whole level attention and iteratively parses the two levels with the proposed encoder-decoder interaction. Further, MaskFormer~\cite{cheng2021per} uncovers the importance of per mask and replaces the per-pixel classification task with a set of binary masks prediction, each associated with a single class.
However, the information flow between class embeddings and predicted masks is one-way, from former to the latter. While much prior knowledge lying on accurate regions of each slice in masks is completely ignored and lost.

\subsection{Multi-stage aggregation}

Aggregating multi-stage features is an important method to improve recognition accuracy. Top-down feature fusion is used in~\cite{pinheiro2016learning, honari2016recombinator}, which aims to optimize low-resolution features by using higher features stage by stage and eliminating the spatial information loss caused by down-sampling. DSSD~\cite{fu2017dssd}, TDM~\cite{shrivastava2016beyond} explored different approaches to improve feature aggregation, \eg employing complex residual connections. FPN~\cite{lin2017feature} proposes a feature pyramid architecture that combines multi-stage features via a top-down pathway and lateral connections. After that, PANet~\cite{liu2018path}, NAS-FPN~\cite{ghiasi2019fpn}, BiFPN~\cite{tan2020efficientdet} and Recursive FPN~\cite{qiao2021detectors} explore the geometric topology of the feature pyramid network to seek the optimum.

In contrast, we consider interacting and transferring semantic information between multi-stage features with the help of highly consistent class embeddings.

\section{Methodology}

In this section, we describe the proposed ECENet in detail.
An overview of the model is shown in Fig.~\ref{fig3}. The feature maps 
encoded from input images are briefly introduced in Section \ref{sec:3-1}. To enhance the discriminability and informativity of features, we contribute a way to rebuild the features from backbone (Section \ref{sec:3-2}). The proposed Explicit Class Extraction (ECE) and Semantics Attention \& Updater (SAU), are described respectively in Section \ref{sec:3-3} and \ref{sec:3-4}. And we conclude our whole model in Section \ref{sec:3-5}.

\subsection{Encoder}
\label{sec:3-1}
Given an input image $I \in \R^{H \times W \times 3}$, it is transformed and reshaped into a sequence of tokens $\mathcal{F}_{0} \in \R^{L \times C}$ where $L = HW/P^2$, $P$ is the patch size and $C$ is the number of channels.
With positional information involved, the token sequence $\mathcal{F}_{0}$ is passed into $4$ stages, each contains several transformer layers. Typically, there are patch merging modules between stages to gradually reduce the resolution of feature maps. After that, we obtain multi-level features at $ \{1/4, 1/8, 1/16, 1/32\}$ of the original image resolution, which are defined as $[\mathcal{F}_{1}, \mathcal{F}_{2}, \mathcal{F}_{3}, \mathcal{F}_{4}]$.
$\mathcal{F}_{i} \in \R^{L_{i} \times C_{i}}$ where $L_{i} = H_{i} \times W_{i}$, represents the $i$th-stage feature whose scale is $1/2^{i+1}$ of the input image, \ie $H_i = H/2^{i + 1}$ and $W_i = W/2^{i + 1}$, $C_{i}$ is the embedding dimensions of stage $i$, $\forall i = 1, 2, 3, 4$.

It is worth noting that we introduce the typical hierarchical transformer backbone as our encoder here.
However, our proposed method is also appropriate for plain ViT or the CNN backbone.

\subsection{Feature reconstruction (FR)}
\label{sec:3-2}

The diversity and redundancy in feature maps is an important characteristic of successful networks~\cite{han2020ghostnet}, but has rarely been investigated, resulting in the unknown and hard-to-control feature representations. We point out that it is actually avoidable and can be improved.

\begin{figure}
	\centering
	\includegraphics[width=0.97\linewidth]{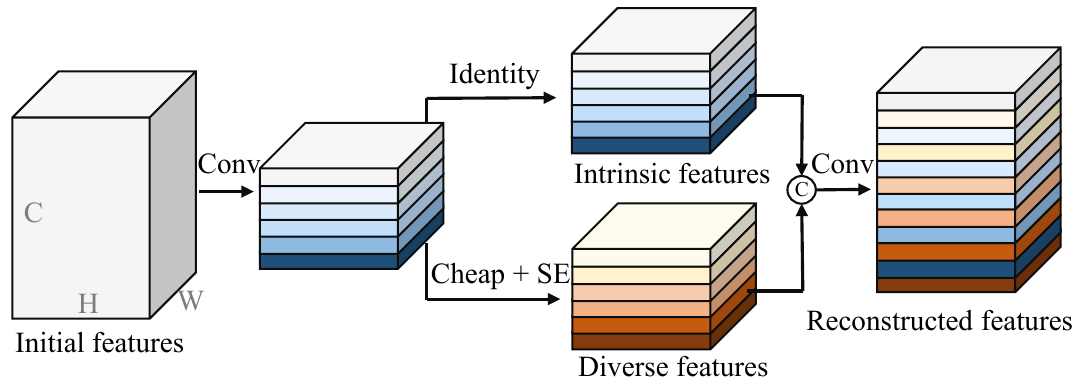}
	\caption{Feature Reconstruction module. It consists of two branches, named as intrinsic features and diverse features. By combining the two branches together, we ensure the \textbf{diversity} and \textbf{redundancy} simultaneously existing in features.}
	\label{fig2}
\end{figure}

To this end, we design a simple yet effective approach to purposefully control the diversity and redundancy of feature maps, dubbed Feature Reconstruction (FR). It can ensure 
better representation capabilities of features,
thus assist in obtaining good-qualified semantic embeddings later.
As observed in Fig.~\ref{fig2}, we assume that the output feature map in each stage, $\mathcal{F}_{i} \in \R^{L_{i} \times C_{i}}$, which can be reshaped as $\mathcal{F}_{i} \in \R^{ H_{i} \times W_{i} \times C_{i}}$, actually 
contains intrinsic features and we extract them using a 1x1 convolution 
followed by a norm layer.
\begin{equation}
	\mathbf{Y^{'}} = \phi(\mathcal{F}_{i}) \in \R^{H_{i} \times W_{i} \times C_{i} // 2}
\end{equation}

\begin{figure*}[t]
	\centering
	\includegraphics[width=0.92\linewidth]{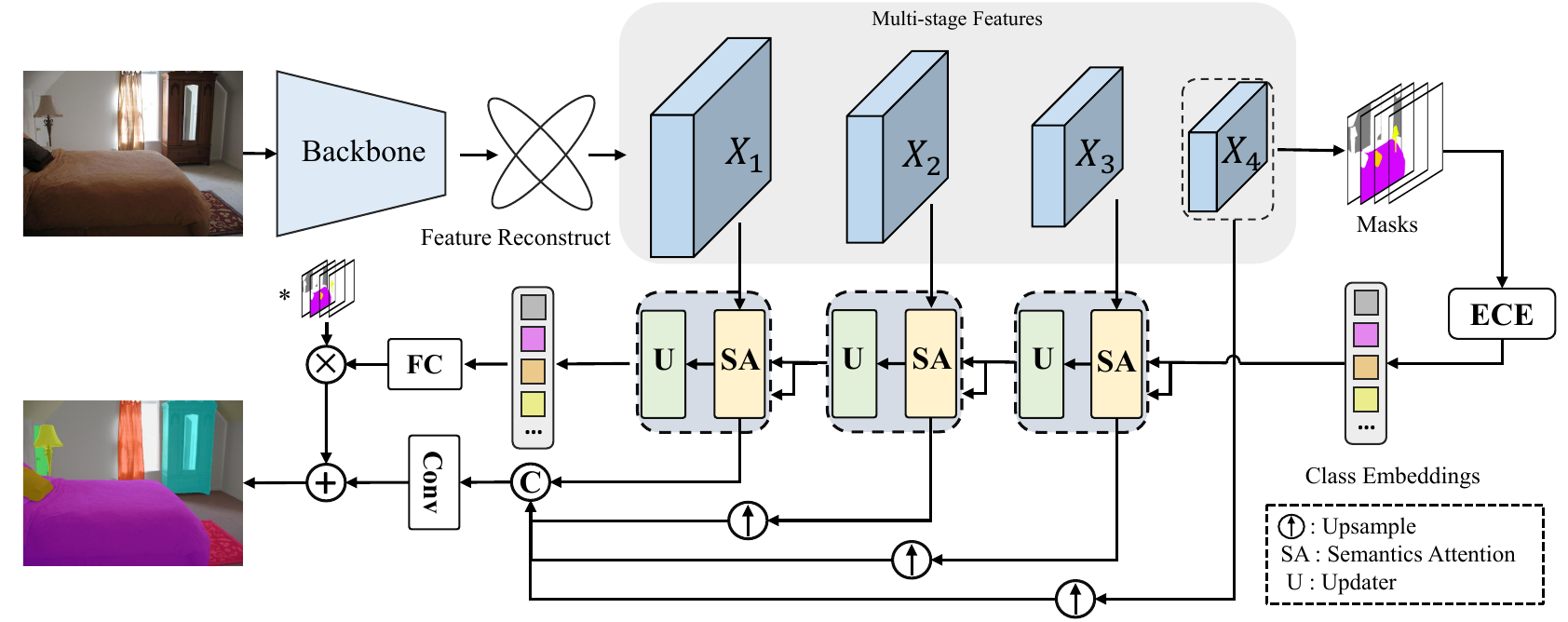}
	\caption{\textbf{The overall structure of ECENet.} It's composed of Feature Reconstruction (FR), Explicit Class Extraction (\ece), Semantics Attention \& Updater (SAU). The FR module ensures the discriminative and informative capability of features from backbone. Then explicit class embeddings are generated from features of final stage by ECE module. After that, we carry out SAU module sequentially to make previous stage features interact with the class embeddings, thus higher-level semantics are transferred gradually. Masks emerge as a byproduct in the attention mechanism and are used to enhance our class embeddings. Finally, the enhanced multi-features are aggregated to get final predictions, assisted by class embeddings and summed masks. `*' means multiple masks.}
	\label{fig3}
\end{figure*}

To further ensure the diversity, we propose another diverse branch, which is realized by cheap linear operations on each intrinsic feature in $\mathbf{Y^{'}}$, following GhostNet~\cite{han2020ghostnet}. However, we argue that this transform is uncontrollable, which indicates that the sensitivity to informative features is low. Thus we apply SE module~\cite{hu2018squeeze} to re-calibrate channel-wise feature responses, ensuring that the useful ones are exploited more, as follows:

\begin{equation}
	\mathbf{Y^{''}} = {\rm SE}({\rm Cheap}(\mathbf{Y^{'}}) \in \R^{H_{i} \times W_{i} \times C_{i} // 2}
\end{equation}

Furthermore, the Softmax function is applied on each channel to exclusively concentrate on the most notable regions. Then all ‘max’ values on the feature maps are summed which aims to ensure that each pixel is uniquely covered, especially for our pixel-level dense prediction tasks.
Notably, we add minus in the summation results, namely diversity loss, $\mathcal{L}_{div}$. The formulation is:
\begin{equation}
	{L}_{div}(\mathbf{Y^{''}})=1 - \frac{1}{\mathcal{C}}\sum_{k=1}^{HW}\max_{j=1,2,\cdots,\mathcal{C}}\frac{e^{\mathbf{Y^{''}}_{j,k}}}{\sum_{k'=1}^{W\!H}e^{\mathbf{Y^{''}}_{j,k'}}}
	\label{equ:ldiv}
\end{equation}
where $\mathcal{C}$ equals to half of the channels. Here we emphasis that in dense prediction tasks, each pixel should be noticed, thus making each slice of feature maps unique is reasonable. This process requires almost no extra computation. And it can improve the performance by a large margin, which is further shown in ablation experiments.

Finally, the intrinsic branch $\mathbf{Y^{'}}$ and diverse branch $\mathbf{Y^{''}}$ are concated together and projected to the initial shape. By this way, we ensure the diversity and redundancy simultaneously existing in each stage's representative feature maps.

\subsection{Explicit class extraction (\ece)}
\label{sec:3-3}
Intuitively, accurate regions on each slice of predicted masks become the most natural description of each category. With this insight, we seek to get explicitly defined class embeddings by taking use of predicted masks. 

Specifically, after feature reconstruction, we unify their channels by a 1x1 convolution and define the unified features as $[\mathcal{X}_{1}, \mathcal{X}_{2}, \mathcal{X}_{3}, \mathcal{X}_{4}]$, where $\mathcal{X}_{i} \in \R^{C \times H_{i} \times W_{i}}$, $\forall i = 1, 2, 3, 4$, each holds \{1/4, 1/8, 1/16, 1/32\} of the original image resolution. $C$ is the unified dimension, usually set to 256.
Then, linear transformations are applied to the feature map $\mathcal{X}_{4}$ from last stage, as presented by Eq.~\eqref{eq:1}.
\begin{equation}
	Mask(\mathcal{X}_{4}) = \phi_2 (\phi_1 (\mathcal{X}_{4}))
	\label{eq:1}
\end{equation} 
where $\phi_1$ and $\phi_2$ are linear transformations implemented by 1 × 1 convolutional layers without activation. $Mask(\mathcal{X}_{4}) \in \R^{N \times H_{4} \times W_{4}}$ becomes the intermediate masks with N equals to the number of classes. 

Next, the Explicit Class Extraction module is employed, as illustrate in Fig.~\ref{fig1}. Specifically, each slice of the masks is divided into spatial bins. 
Then we conduct parameter-free pooling in each spatial bin to harvest sub-region representations.

But this operation is non-trivial. We aim to cover class information on different sized areas. Inspired by~\cite{he2015spatial}, we use spatial pyramid pooling to maintain both local and global information. Intuitively, the biggest divided number of a side ought to depend on datasets. Thus the maximum pooling size is set to be proportional to $\sqrt{N}$, \ie $\alpha \sqrt{N}$, where $N$ is the total number of classes. Ratio $\alpha$ is adjusted to different datasets. Then a single linear projection just follows to convert the channels to unified dimension $C$ and obtain the explicit class embeddings $\mathcal{G} \in \R^{N \times C}$. The whole process can be summarized in Eq.~\eqref{eq:2}.
\begin{equation}
	\mathcal{G} = \psi ({\rm Pooling} (Mask(\mathcal{X}_{4})))
	\label{eq:2}
\end{equation}
where $\psi$ is a simple linear transformation. Through ECE module, we bridge the gap between segmentation masks and class embeddings, allow the information flow reversely and obtain the explicit class embeddings instead of random initialized blindly, which completes the last piece of the puzzle in recent segmentation tasks.

\subsection{Semantics attention \& updater (SAU)}
\label{sec:3-4}
This part integrates a Semantics attention module used for interacting between image features and explicit class embeddings, and an update head which refreshes the class representations using the newly obtained ones.

\begin{figure}
	\centering
	\includegraphics[width=0.85\linewidth]{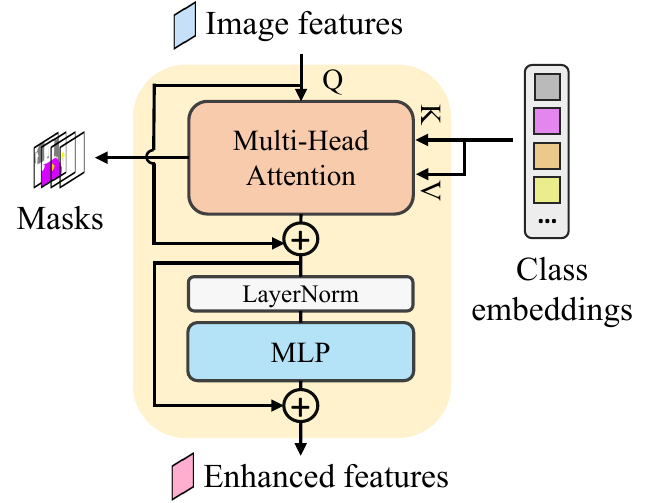}
	\caption{Semantics Attention module (SA). Image features are strengthened during interacting with explicit class embeddings from higher stages. Note that the LayerNorm applied on image and class embeddings is omited for simplicity.}
	\label{fig4}
\end{figure}

\textbf{Semantics Attention.} As illustrated in Fig.~\ref{fig4}, the Semantics Attention module consists of a multi-head attention (MSA) block and a MLP block with residual connections added after every block.
Specifically, it takes earlier features $\mathcal{X}_{i-1}$ as queries and class embeddings $\mathcal{G} \in \R^{N \times C}$ as keys and values. Firstly, linear transformations are applied to form query ($Q$), key ($K$) and value ($V$), as presented by Eq.~\eqref{eq:3}.
\begin{equation}
	\begin{split}
	Q & = \phi_{q} (\mathcal{X}_{i-1}) \in \R^{L_{i-1} \times C}, \\
	K & = \phi_{k} (\mathcal{G}) \in \R^{N \times C}, \\
	V & = \phi_{v} (\mathcal{G}) \in \R^{N \times C}.
	\label{eq:3}
	\end{split}
\end{equation}
Following the scaled dot-product attention mechanism, $Q$ and $K$ are interacted to measure Class-Feature similarity. The similarity map and attention map are calculated as following:
\begin{equation} 
	\begin{split}
		S(Q, K)& = \frac{QK^T}{\sqrt{d_{k}}}, \\
		{\rm MSA}(\mathcal{G}, \mathcal{X}_{i-1}) & = 
		{\rm Softmax}(S(Q, K))V.
	\end{split}
\end{equation}
where $\sqrt{d_{k}}$ serves as a scaling factor while $d_{k}$ equals to the dimension of keys, $S(Q, K) \in \R^{L_{i-1}  \times N}$, and ${\rm MSA}(\mathcal{G}$, $\mathcal{X}_{i-1}) \in \R^{L_{i-1} \times C}$. 
With the guidance of clearly defined class embeddings, the regions belonging to the same category tend to group together and return strengthened representative features. 

Then a MLP block of three layers is applied. Note that, we follow~\cite{xie2021segformer} to employ a $3 \times 3$ depth-wise convolution which considers the effects of zero padding to leak position information. LayerNorm (LN) is applied before every block.

\textbf{Class Updater.}
The clearly defined class embeddings initially emerge from features of final stage. However, in segmentation tasks, models need to recognize regions that vary in scale, which call for multi-scale spatial information enhancement. Same goes for our explicit class embeddings.
Fortunately, the accurate masks appear in an inner way, as a byproduct where $Q$ and $K$ interact to measure Class-Feature similarity. Inspired by~\cite{zhang2022segvit}, we reshape and utilize this similarity map as new generated masks, where $S(Q, K) \in \R^{L_{i-1}  \times N}$, to $Mask(\mathcal{X}_{i-1}, \mathcal{G}) \in \R^{N \times H_{i-1} \times W_{i-1}}$. 
After that, we employ Explicit-Class-Extraction (\ece) module to get newly obtained class embeddings $\hat{\mathcal{G}}$:
\begin{equation}
	\hat{\mathcal{G}} = {\rm ECE}(Mask(\mathcal{X}_{i-1}, \mathcal{G}))
\end{equation}
Then gated mechanism is applied to further refresh previous class embeddings. Specifically, we first fuse the new class representations $\hat{\mathcal{G}}$ with previous ones as
\begin{equation}
	F_u = \phi_{3}(\hat{\mathcal{G}}) \odot \phi_{4}(\mathcal{G})
\end{equation}
where $F_u \in \R^{N \times C}$, $\phi_{3}$ and $\phi_{4}$ are linear transformations. Then an updating gate, $U$ is learned and multiplied with the projected $\hat{\mathcal{G}}$.
Finally, we add this with the previous $\mathcal{G}$ to get the updated class embeddings, as follows:
\begin{equation}
	\begin{split}
		U & = \sigma(\psi_{1}(F_u)) \\
		\mathcal{G} & = U \odot \psi_{2}(\hat{\mathcal{G}}) + \mathcal{G}
	\end{split}
\end{equation}
where $\sigma$ is the Sigmoid function and $\psi_{1}$, $\psi_{2}$ are fully connected (FC) layers followed by LayerNorm (LN), as in~\cite{zhang2021k}.
This Semantics Attention \& Updater module repeats progressively on the remaining multi-stage features, resulting highly informative class embeddings. Furthermore, the representative image features are enhanced during Class-Feature interaction, eliminating semantic gap between different layers.

\subsection{The ECENet structure}
\label{sec:3-5}
As illustrated in Fig.~\ref{fig3}, we get reconstructed features $[\mathbf{Y}_{1}, \mathbf{Y}_{2}, \mathbf{Y}_{3}, \mathbf{Y}_{4}]$ from FR module using multi-stage features and unify their channels to get $[\mathcal{X}_{1}, \mathcal{X}_{2}, \mathcal{X}_{3}, \mathcal{X}_{4}]$. Then explicit class embeddings is extracted from last feature $\mathcal{X}_{4}$.
Previous stage features interact with the class embeddings and higher-level semantics are transferred gradually. Masks emerge as a byproduct in the attention mechanism and are used to enhance our class embeddings.

Finally, the enhanced multi-stage features are up-sampled using ghost feature~\cite{han2020ghostnet} and pixel shuffle~\cite{shi2016real}. Then all features are concated together and pass through a 1x1 convolution to get the final prediction. As in DETR~\cite{carion2020end} and SegViT~\cite{zhang2022segvit}, we apply a linear transformation followed by a Softmax activation to the final class embeddings to get class probability predictions. Then it multiplies with summed masks, and added on to enhance the final prediction.  
Formally, the loss function can be formulated as:
\begin{equation}
	\mathcal{L}_{overall} = \mathcal{L}_{cls} +\mathcal{L}_{mask} + \lambda_{div}\mathcal{L}_{div}
\end{equation}
where
\begin{equation}
	\mathcal{L}_{mask} = \lambda_{focal}\mathcal{L}_{focal} + \lambda_{dice}\mathcal{L}_{dice}
\end{equation}

The classification loss ($\mathcal{L}_{cls}$) is implemented by Cross-Entropy loss and the masks are summed together, supervised by the mask loss ($\mathcal{L}_{mask}$) which is a linear combination of a focal loss and a dice loss multiplied by hyper-parameters $\lambda_{focal}$ and $\lambda_{dice}$ respectively as in DETR~\cite{carion2020end}. The $\mathcal{L}_{div}$ is applied on each stage's `diverse' branch features. The experiments in next section show that this design is efficient and works well.

\textbf{Different from Existing Methods.} 
Though this mask-to-class transform is the converse of traditional prediction process, there are still some works that vaguely realize the importance of this. In CNN networks, ACFNet~\cite{zhang2019acfnet} tried to perform matrix multiplication between feature maps $F \in \R ^ {C \times H \times W}$ and coarse mask $P \in \R ^ {N \times H \times W}$. However, it faces challenges if only a pure mask is given.
Instead, we emphasize the predicted mask itself and reveal the underlying mechanism.
The accurate regions on predicted masks present each category independently, without the assistance of feature maps $F$. Actually, if extra feature maps are brought in to get class embeddings, it actually confuses the purely meaningful region representations in masks.

Further, our method not only improves the efficiency, but also has better interpretability by incorporating class information into embeddings, as shown later in Section \ref{ablation_sec}. Basically, both our method and Mask2Former~\cite{cheng2022masked} recognize and utilize the value in masks which localize regions, but with different methods. Mask2Former~\cite{cheng2022masked} uses 100+ queries and masked-attention that aims to focus on specific regions. While we have new insights on masks and transfer these
into Explicit Class Embeddings, which is more explainable. And this idea is well suited to other frameworks.

Besides, we want to emphasize that though good performance is achieved in some works, \eg K-Net~\cite{zhang2021k} which introduces dynamic class kernels, however, the boosting in performance actually is lying in the usage of supervised masks to polish the learned kernels gradually.

\section{Experiments}

\subsection{Datasets}

\textbf{ADE20K} 
is a scene parsing dataset covering $150$ fine-grained semantic concepts, consisting of $20,210$ images as the training set and $2,000$ images as the validation set.

\textbf{PASCAL-Context} 
contains $4,996$ and $5,104$ images for training and validation respectively. Following previous works, we evaluate on the most frequent $59$ classes.

\textbf{Cityscapes}
is a driving-scene dataset densely annotates $19$ object categories in images. It contains $5,000$ finely annotated images, split into $2,975/500/1,525$ for training, validation and testing respectively.

\subsection{Implementation details}
\textbf{Training settings}
We mainly use the Swin Transformer as the backbone. Specifically, we provide results primarily on its `Large' variation and use its `Base' variation for most ablation studies. Our experiments are based on MMSegmentation~\cite{mmseg2020} and follow the commonly used training settings.
During training, we use AdamW as the optimizer with a total iteration of $160k$, $80k$ and $160k$ for ADE20K, PASCAL-Context and Cityscapes respectively. The batch size is set to 16 except that we use batch size 8 for Cityscapes.
Ratio $\alpha$ is empirically set to $1$, $\sqrt{2}$, $3$ for ADE20K, PASCAL-Context and Cityscapes respectively.
We employ data augmentation sequentially via random resize with the ration between $0.5$ and $2.0$, random horizontal flipping, and random cropping ($640\times 640$ for ADE20K, $480 \times 480$ for PASCAL-Context and $1024\times 1024$ for Cityscapes).

\textbf{Evaluation metric.}
We use the mean intersection over union (mIoU) to evaluate the segmentation performance. All reported mIoU scores are in a percentage format. `ss' means single-scale testing and `ms' means test time augmentation with multi-scaled $(0.5, 0.75, 1.0, 1.25, 1.5, 1.75)$ inputs. All reported number of parameters (Params) and computational costs in GFLOPs are measured using the fvcore~\footnote{\url{https://github.com/facebookresearch/fvcore}} library.

\subsection{Comparison with State-of-the-Arts}

\begin{table*}
	\begin{center}
		\resizebox{0.75\linewidth}{!}{
			\begin{tabular}{rlccccc}
				\toprule[1.3pt]
				Method & Backbone & Crop Size & GFLOPs & \#param. & mIoU (ss) &  mIoU(ms) \\
				\midrule
        		Zhou et al.~\cite{zhou2022rethinking} & ResNet-101  & $512 \times 512$ & - & 68.5M  & 41.1 & - \\
				PSPNet~\cite{zhao2017pyramid} & ResNet-101  & $512 \times 512$ & 257 & 65.7M  & 44.4 & 45.4 \\
				UPerNet~\cite{xiao2018unified} & ResNet-101 & $512 \times 512$  & 258 & 85.5M  & 43.8 & 44.9 \\
                    \midrule
                    \enet\ (Ours) & ResNet-101  & $512 \times 512$ & 293.1 & 60.2M  & \textbf{45.3} & \textbf{46.8} \\
                    \midrule
                    \midrule
				DPT~\cite{ranftl2021vision} &ViT-Base& $512 \times 512$ & 219.8 & - &47.2 & 47.9 \\
				StructToken-SSE \cite{lin2022structtoken}  & ViT-Base  & $512 \times 512$& \textgreater150 & 142M & 50.9 & 51.8\\
                    UPerNet + SwinT~\cite{liu2021swin} & SwinT-Base$\dagger$  & $640 \times 640$& 471 & 121.4M & - & 51.6\\
				\midrule
				\enet\ (Ours) &SwinT-Base$\dagger$ & $640 \times 640$ & 243.3 & 96.7M & \textbf{53.4} & \textbf{54.2} \\
				\midrule
				\midrule
				
				SETR-PUP~\cite{zheng2021rethinking} & ViT-Large & $640 \times 640$ & 711 & 308M & 48.2 & 50.0 \\
				Segmenter~\cite{strudel2021segmenter} & ViT-Large  & $640 \times 640$  & 672 & 333M& 51.7 & 53.6 \\
				StructToken-CSE~\cite{lin2022structtoken} & ViT-Large$\dagger$  & $640 \times 640$ & \textgreater700 & 350M  & 52.8 & 54.2 \\
				UPerNet + ViT-Adapter~\cite{chen2022vision} & ViT-Large$\dagger$  & $640 \times 640$ & - & 364M  & 53.4 & 54.4 \\
				UPerNet + SwinT~\cite{liu2021swin} & SwinT-Large$\dagger$  & $640 \times 640$ & 647 & 234M  & - & 53.5 \\
				UperNet + KNet~\cite{zhang2021k} & SwinT-Large$\dagger$  & $640 \times 640$ & 659 & 245M & - & 54.3 \\
				SenFormer~\cite{bousselham2021efficient} & SwinT-Large$\dagger$  & $640 \times 640$ & 546 & 233M & 53.1 & 54.2 \\
				\midrule
				\enet\ (Ours) & SwinT-Large$\dagger$ & $640 \times 640$ & 425.7 & 208.3M & \textbf{54.1} & \textbf{55.2}\\
				\bottomrule[1.3pt]
			\end{tabular}
		}
	\end{center}
	\caption{Experiment results on the ADE20K \texttt{val.}\  split. `$\dagger$' means the model's weight are pretrained on
 ImageNet-22K. The GFLOPs is measured at single-scale inference with the given crop size. Note that we also use dilated ResNet-101 as backbone.}
	\label{tab:1}
\end{table*}

\textbf{Results on ADE20K.}
Table \ref{tab:1} reports comparison with the state-of-the-art methods on ADE20K validation set. 
Our \enet\ outperforms other counterparts by a large margin on diverse backbones with relatively small amount of parameters. With the SwinT-Large backbone, our \enet\ achieves $55.2\%$ mIoU using only $208.3$M paramenters and $425.7$ GFLOPs. It is $1.0\%$ mIoU better than the recent SenFormer~\cite{bousselham2021efficient} using the same backbone. Noting that, the parameters and computational cost of our method is much less than others. Our method based on SwinT-Base backbone achieves $53.4\%$ mIoU under single-scale inference, which is similar to SenFormer~\cite{bousselham2021efficient} ($53.1\%$ mIoU) based on SwinT-Large and UPerNet + ViT-Adapter~\cite{chen2022vision} ($53.4\%$ mIoU) based on ViT-Large, but with much less parameters ($96.7$M vs. $364$M) and computational cost ($243.3$ GFLOPs vs. $546$ GFLOPs).

\textbf{Results on PASCAL-Context.}
Table \ref{tab:2} shows the result on PASCAL-Context dataset. We follow SenFormer~\cite{bousselham2021efficient} to evaluate our method and report the results  under $59$ classes. Our \enet\ achieves mIoU $65.9\%$, which outperforms the recent SenFormer~\cite{bousselham2021efficient} with the same SwinT-Large backbone by $1.4\%$ mIoU, and achieves new state-of-the-art performance. Compared with SegViT~\cite{zhang2022segvit}, our \enet\ outperforms it by $0.6\%$ mIoU  with much less computational cost (reduced $25\%$ GFLOPs).

\begin{table}
	\begin{center}
	\resizebox{\linewidth}{!}{
	\begin{tabular}{rlcc}
		\toprule[1.3pt]
		Method & Backbone & GFLOPs &  mIoU(ms) \\
		\midrule
		PSPNet~\cite{zhao2017pyramid} & ResNet-101 & 157 & 47.8 \\
		EncNet~\cite{zhang2018context} & ResNet-101 & 192.1 & 52.6 \\
		HRNetv2~\cite{wang2020deep} & HRNetv2-W48 & 82.7 & 54.0 \\
		NRD~\cite{zhang2021dynamic} & ResNet-101 & 42.9 & 54.1 \\
		CAA~\cite{huang2022channelized} & ResNet-101 & - & 55.0 \\
		SegViT (Shrunk)~\cite{zhang2022segvit} & ViT-Large & 186.9 & 63.7 \\
		SegViT~\cite{zhang2022segvit} & ViT-Large & 321.6 & 65.3 \\
		UPerNet + CAR~\cite{huang2022car} & SwinT-Large$\dagger$ & - & 59.0 \\
		SenFormer~\cite{bousselham2021efficient} & SwinT-Large$\dagger$ & \textgreater300 & 64.5 \\
		
		\midrule
		\enet\ (Ours) & SwinT-Large$\dagger$ & 241.9 & \textbf{65.9}\\
		\bottomrule[1.3pt]
	\end{tabular}
	}
	\end{center}
	\caption{Experiment results on PASCAL-Context validation set with multi-scale inference. `$\dagger$' 
 means the model's weight are pretrained on
 ImageNet-22K. The GFLOPs is measured at single-scale inference with a crop size of $480 \times 480$. }
	\label{tab:2}
\end{table}

\textbf{Results on Cityscapes.} 
Table \ref{tab:3} shows the result on Cityscapes validation set. Our method reaches mIoU $84.5\%$, surpassing Mask2Former~\cite{cheng2022masked} by $0.2\%$ mIoU, which is very competitive with the previous works.

\begin{table}
	\begin{center}
	\resizebox{0.86\linewidth}{!}{
	\begin{tabular}{rlcc}
		\toprule[1.3pt]
		Method & Backbone &  mIoU(ms) \\
		\midrule
		PSPNet~\cite{zhao2017pyramid} & ResNet-101  & 78.5 \\
		DeepLabv3+~\cite{chen2018encoder} & Xception-71 & 79.6 \\
		CCNet~\cite{huang2019ccnet} & ResNet-101  & 81.3 \\
		CANet~\cite{tang2022compensating} & ResNet-101 & 81.9 \\
		AlignSeg~\cite{huang2021alignseg} & ResNet-101 & 82.4 \\
		P-DeepLab~\cite{cheng2020panoptic} & Xception-71 & 81.5 \\
		SegFormer~\cite{xie2021segformer} & MiT-B5 & 84.0 \\
		SETR-PUP~\cite{zheng2021rethinking} & ViT-Large & 82.2 \\
		Segmenter~\cite{strudel2021segmenter} & ViT-Large & 81.3 \\
		StructToken-PWE~\cite{lin2022structtoken} & ViT-Large & 82.1 \\
		Mask2Former~\cite{cheng2022masked} & SwinT-Large$\dagger$ & 84.3 \\
		
		\midrule
		\enet\ (Ours) & SwinT-Large$\dagger$  & \textbf{84.5}\\
		\bottomrule[1.3pt]
	\end{tabular}
	}
	\end{center}
	\caption{Experiment results on Cityscapes validation set with multi-scale inference. `$\dagger$' 
 means the model's weight are pretrained on
 ImageNet-22K. }
	\label{tab:3}
\end{table}

\begin{figure*}[ht]
	\centering
	\includegraphics[width=1.0\linewidth]{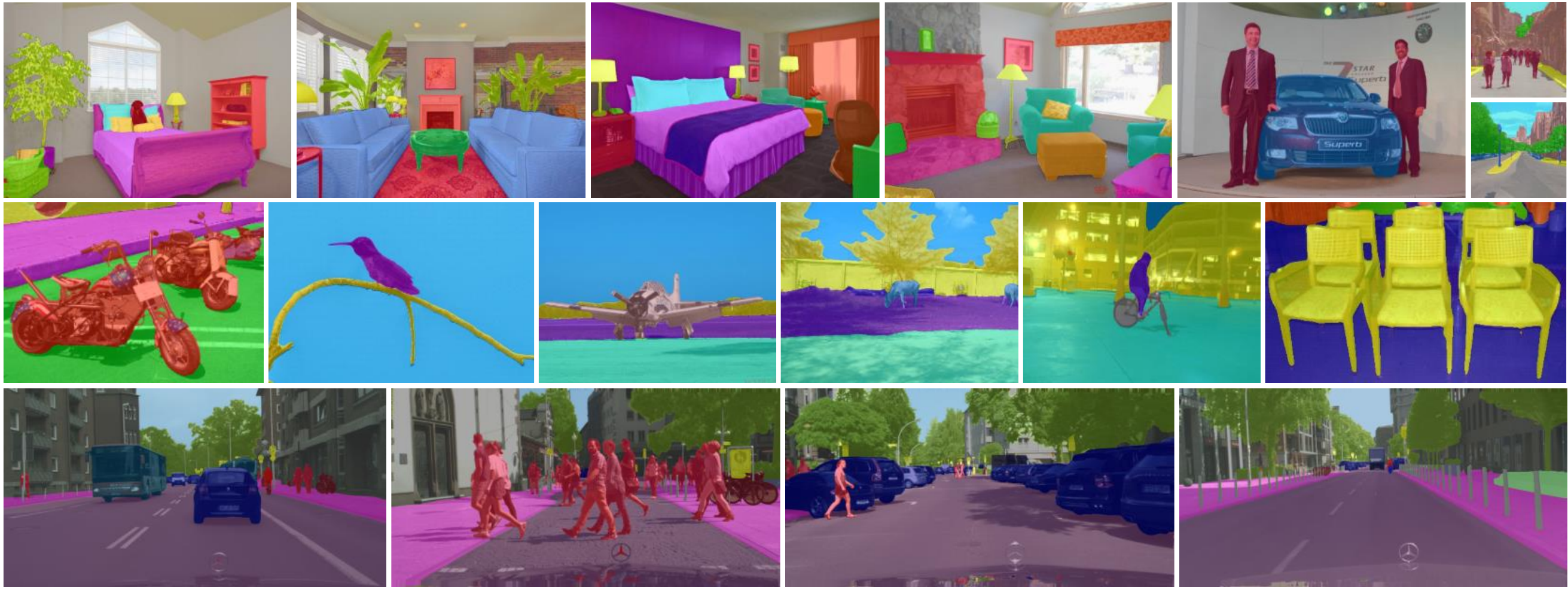}
	\caption{Competitive segmentation results on the ADE20K, PASCAL-Context and Cityscapes validation set. }
	\label{fig5}
\end{figure*}

\subsection{Ablation study} \label{ablation_sec}

In this section, we conduct experiments on ADE20K dataset with SwinT-Base backbone to show the effectiveness of our proposed ECENet method.

\textbf{Ablation of the components in ECENet}
Table \ref{tab:4} shows the effect of different components in ECENet.
Dynamic tokens means that the class embeddings are updated by masks which is a byproduct of attention. We can see that Feature Reconstruction is capable of providing $0.6\%$ mIoU of performance promotion. It's more beneficial to make use of the diversity loss, $\mathcal{L}_{div}$ to make sure the diversity and redundancy in feature maps before interaction.
\begin{table}
	\begin{center}
	\resizebox{0.97\linewidth}{!}{
	\begin{tabular}{ccccc}
		\toprule[1.3pt]
		Dynamic tokens & FR(wo $\mathcal{L}_{div}$)  & $\mathcal{L}_{div}$ & mIoU (ss) & GFLOPs \\
		\midrule
	    \checkmark & & & 52.5 & 240.7 \\
		\checkmark & \checkmark &  & 53.1 & 243.3 \\
		\checkmark & \checkmark & \checkmark & 53.4 & 243.3 \\
		
		\bottomrule[1.3pt]
	\end{tabular}
	}
	\end{center}
	\caption{Ablation results on the components in ECENet. The experiment is carried on ADE20K dataset with SwinT-Base backbone. The crop size is  $640 \times 640$. `FR': Feature Reconstruction. }
	\label{tab:4}
\end{table}

\textbf{Ablation of loss coefficient}
Table \ref{tab:5} shows the ablation of loss coefficient $\lambda_{div}$. The adopted choice of loss coefficient $\lambda_{div}$ is purely result-driven. We compare the model performance under 3 levels of loss coefficient, 0.1, 0.2, 0.5. Experiments show that our ECENet achieves the optimal $53.4\%$ mIoU when $\lambda_{div} = 0.2$. We thus use this setting for all experiments.

\begin{table}
	\begin{center}
	\resizebox{0.35\linewidth}{!}{
	\begin{tabular}{cc}
		\toprule[1.3pt]
		$\lambda_{div}$ & mIoU (ss) \\
		\midrule
		0.1 & 53.25  \\
		0.2 & \textbf{53.40}  \\
		0.5 & 52.65  \\
		
		\bottomrule[1.3pt]
	\end{tabular}
}
	\end{center}
	\caption{Ablation results on loss coefficient $\lambda_{div}$. The experiments use the SwinT-Base backbone and are carried out on ADE20K dataset.}
	\label{tab:5}
\end{table}

\begin{table}
	\begin{center}
	\resizebox{0.45\linewidth}{!}{
	\begin{tabular}{cc}
		\toprule[1.3pt]
		Updater & mIoU (ss) \\
		\midrule
		plus & 52.7 \\
		gated &  \textbf{53.4} \\
		
		\bottomrule[1.3pt]
	\end{tabular}
}
\end{center}
\caption{Ablation results on the updating method of the Updater on ADE20K dataset using SwinT-Base as backbone. }
\label{tab:6}
\end{table}

\textbf{Ablation of the updating method.}
To demonstrate the effectiveness of our class updater, we conduct comparison between the naive plus and gated operation, observed in Table \ref{tab:6} . Experiments show that our gated operation surpasses the naive plus by $0.7\%$ mIoU. 

\textbf{Applying to other methods.}
Since we get the insight that supervised masks could serve as the best tool to get class embeddings, we wish to study the effect of our explicit class embeddings obtained from masks when cooperating with randomly-initialized ones. Thus we conduct ablation studies on the recent SegViT~\cite{zhang2022segvit}. As shown in Table \ref{tab:7}, the application of our gated-updating on the original randomly initialized class embeddings could also provide a performance boost by a large margin of $0.6\%$ mIoU, which shows our potential on different structures and methods. Besides, the gain on the computational cost is negligible.

\begin{table}[h]
	\begin{center}
	\begin{tabular}{ccc}
		\toprule[1.3pt]
		Method & mIoU (ss) & GFLOPs\\
		\midrule
		original & 51.3 & 120.9 \\
		gated-updating &  \textbf{51.9} & 121.2  \\
		
		\bottomrule[1.3pt]
	\end{tabular}
\end{center}
\caption{Verification of our gated-updating strategy using SegViT on ADE20K. The backbone is ViT-Base. The GFLOPs are measured at single-scale inference with a crop size of $512 \times 512$. }
\label{tab:7}
\end{table}

\subsection{Visualization}
The visualization results of our ECENet with SwinT-Large as the backbone are shown in Fig.~\ref{fig5}. Notably, our method can produce satisfactory segmentation results in various indoor and outdoor scenes on ADE20K, PASCAL-Context and Cityscapes validation set, especially at the edges of objects, \eg trees, leaves and chairs. 
This demonstrates that our model can achieve extraordinary promotion in exploiting implications of class embeddings. There is great potential in exploring the flow of information from masks to class embeddings and we can gain a deeper understanding of the category semantics learned by the model.
\section{Conclusion}

This work utilizes explicit class embeddings to boost the performance of semantic segmentation. In contrast to the existing randomly initialized class embeddings which are content-ignored and implicit, we generate them with the coarsely predicted segmentation masks, which makes them consistent and meaningful initially. Furthermore, with the guidance of clearly defined class embeddings, the regions in feature maps which belong to the same category tend to group together and return strengthened representative features. The explicit class embeddings also alleviate the semantic gap between different layers. Besides, we also propose a novel Feature Reconstruction module that ensure the discriminability and informativity of features from backbone. Our method achieves the new state-of-the-art performance with less computational cost on PASCAL-Context.
Notably, we believe this can spark more interest in revealing the true meanings behind the category semantics.

\noindent \textbf{Acknowledgment}
This research was supported by Huawei Noah’s Ark Lab. We gratefully acknowledge the support of MindSpore, CANN (Compute Architecture for Neural Networks) and Ascend AI Processor used for this research.
In particular, we would also like to thank Yifan Liu, Ning Ding and Bowen Zhang for the help and suggestions at the early stage.

{\small
\bibliographystyle{ieee_fullname}
\bibliography{egbib}
}

\end{document}